# A Classification of 3R Orthogonal Manipulators by the Topology of their Workspace

Maher Baili, Philippe Wenger and Damien Chablat
Institut de Recherche en Communications et Cybernétique de Nantes, UMR C.N.R.S. 6597
1, rue de la Noë, BP 92101, 44321 Nantes Cedex 03 France
Maher.Baili@irccyn.ec-nantes.fr

*Abstract*— A classification of a family of 3-revolute (3R) positioning manipulators is established. This classification is based on the topology of their workspace. The workspace is characterized in a half-cross section by the singular curves. The workspace topology is defined by the number of cusps and nodes that appear on these singular curves. The design parameters space is shown to be divided into nine domains of distinct workspace topologies, in which all manipulators have similar global kinematic properties. Each separating surface is given as an explicit expression in the DH-parameters.

## I. Introduction

A positioning manipulator may be used as such for positioning tasks in the Cartesian space or as the regional structure of a 6R manipulator with spherical wrist. Most industrial regional structures have the same kinematic architecture, namely, a vertical revolute joint followed by two parallel joints, like the Puma. Such manipulators are always *noncuspidal* (i.e. must meet a singularity to change their posture) and they have four inverse kinematic solutions (IKS) for all points in their workspace (assuming unlimited joints). This paper focuses on alternative manipulator designs, namely, positioning 3R manipulators with orthogonal joint axes (orthogonal manipulators). Orthogonal manipulators may have different global kinematic properties according to their link lengths and joint offsets. They may be *cuspidal*, that is, they can change their posture without meeting a singularity [1, 2]. In 1998, ABB-Robotics launched the IRB 6400C, a 6R manipulator to be used in the car industry and designed to minimize the swept volume. The only difference with the Puma was the permutation of the first two link axes, resulting in a manipulator with all its joint axes orthogonal, and *cuspidal*. Commercialization of the IRB 6400C was finally stopped one year later. Exact reason is beyond the knowledge of the authors but the cuspidal behavior is likeable to have disappointed the end users. Cuspidal robots were unknown before 1988 [3], when a list of conditions for a manipulator to be noncuspidal was provided [4, 5]. This list includes simplifying geometric conditions like parallel and intersecting joint axes [4] but also nonintuitive conditions [5]. A general necessary and sufficient condition for a 3-DOF manipulator to be cuspidal was established in [6], namely, the existence of at least one point in the workspace where the inverse kinematics admits three equal solutions. The word "cuspidal manipulator" was defined in accordance to this condition because a point with three equal IKS forms a cusp in a cross section of the workspace [4, 7]. The categorization of all generic 3R manipulators was established in [8] based on the homotopy class of the singular curves in the joint space. [9] proposed a procedure to take into account the cuspidality property in the design process of new manipulators. More recently, [10] applied efficient algebraic tools to the classification of 3R orthogonal manipulators with no offset on their last joint. Five surfaces were found to divide the parameters space into 105 cells where the manipulators have the same number of cusps in their workspace. The equations of these five surfaces were derived as polynomials in the DH-parameters using Groebner Bases. A kinematic interpretation of this theoretical work was conducted in [11] : the authors analyzed general kinematic properties of one representative manipulator in each cell. Only five different cases were found to exist. However, the classification in [11] did not provide the equations of the separating surfaces in the parameters space for the five cells associated with the five cases found. On the other hand, [11] did not take into account the occurrence of nodes, which play an important role for analyzing the number of IKS in the workspace.

The purpose of this work is to classify a family of 3R positining manipulators according to the topology of their workspace, which is defined by the number of cusps and nodes that appear on the singular curves. The design parameters space is shown to be divided into nine domains of distinct workspace topologies, in which all manipulators have similar global kinematic properties. This study is of interest for the design of new manipulators.

The rest of this article is organized as follows. Next section presents the manipulators under study and recalls some preliminary results. The classification is established in section III. Section IV synthesizes the results and section V concludes this paper.

## II. PRELIMINARIES

### A. Manipulators under study

The manipulators studied in this paper are orthogonal with their last joint offset equal to zero. The remaining lengths parameters are referred to as $d_2$, $d_3$, $d_4$, and $r_2$ while the angle parameters $\alpha_2$ and $\alpha_3$ are set to $-90°$ and $90°$, respectively. The

This work was supported in part by C.N.R.S. MathStic program "Cuspidal robots and triple roots".

three joint variables are referred to as $\theta_1$, $\theta_2$ and $\theta_3$, respectively. They will be assumed unlimited in this study. Figure 1 shows the kinematic architecture of the manipulators under study in the zero configuration. The position of the end-tip (or wrist center) is defined by the three Cartesian coordinates $x$, $y$ and $z$ of the operation point $P$ with respect to a reference frame (O, **x**, **y**, **z**) attached to the manipulator base as shown in Fig. 1.

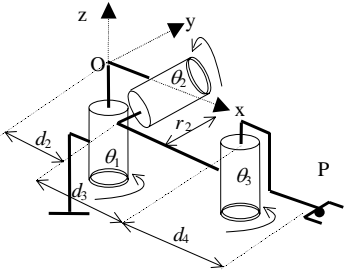

Figure 1. Orthogonal manipulators under study.

### B. Singularities and aspects

The determinant of the Jacobian matrix of the orthogonal manipulators under study is

$$\det(\mathbf{J}) = (d_3 + c_3 d_4)(s_3 d_2 + c_2(s_3 d_3 - c_3 r_2)) \quad (1)$$

where $c_i = \cos(\theta_i)$ and $s_i = \sin(\theta_i)$. A singularity occurs when $\det(\mathbf{J}) = 0$. Since the singularities are independent of $\theta_1$, the contour plot of $\det(\mathbf{J}) = 0$ can be displayed in $-\pi \leq \theta_2 < \pi, -\pi \leq \theta_3 < \pi$ where they form a set of curves. If $d_3 > d_4$, the first factor of $\det(\mathbf{J})$ cannot vanish and the singularities form two distinct curves $S_1$ and $S_2$ in the joint space [12]. $S_1$ and $S_2$ divide the joint space into two singularity-free open sets $A_1$ and $A_2$ called *aspects* [1]. The singularities can be also displayed in the Cartesian space [13, 14]. Thanks to their symmetry about the first joint axis, a 2-dimensional representation in a half cross-section of the workspace is sufficient. The singularities form two disjoint sets of curves in the workspace. These two sets define the internal boundary $WS_1$ and the external boundary $WS_2$, respectively, with $WS_1 = f(S_1)$ and $WS_2 = f(S_2)$. Fig. 2 (left) shows the singularity curves when $d_2 = 1$, $d_3 = 2$, $d_4 = 1.5$ and $r_2 = 1$. For this manipulator, the internal boundary $WS_1$ has four cusp points. It divides the workspace into one region with two IKS (the outer region) and one region with four IKS (the inner region).

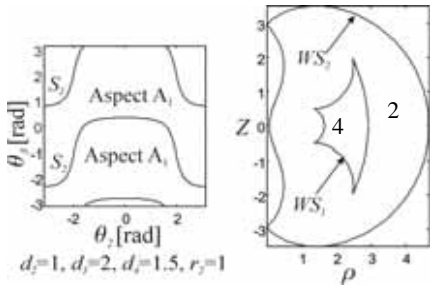

Figure 2. Singularity curves in joint space (left) and workspace (right, number of IKS in each region is indicated).

If $d_3 \leq d_4$, the operation point can meet the second joint axis whenever $\theta_3 = \pm \arccos(-d_3/d_4)$ and two horizontal lines appear, which may intersect $S_1$ and $S_2$ depending on $d_2$, $d_3$, $d_4$ and $r_2$ [12]. The number of aspect depends on these intersections. Note that if $d_3 < d_4$, no additional curve appears in the workspace cross-section but only two points where the operation point meets the second joint axis and the manipulator has an infinite number of IKS. Fig. 3 shows the singularity curves when $d_2 = 1$, $d_3 = 3$, $d_4 = 4$ and $r_2 = 2$. The singular line defined by $\theta_3 = +\arccos(-d_3/d_4)$ maps onto one singular point in the workspace cross-section, which is located at the self-intersection of the internal singular boundary. The remaining singular line $\theta_3 = -\arccos(-d_3/d_4)$ maps onto an isolated singular point in the workspace. The workspace topology of this manipulator features two cusps and three nodes, two regions with two IKS and two regions with four IKS. In the following section, the complete classification is established.

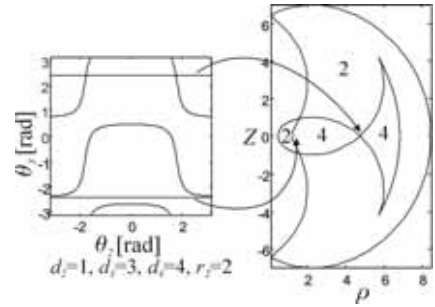

Figure 3. Singularity curves when $d_3 < d_4$. The two horizontal singular lines maps onto isolated singular points in the workspace.

### III. WORKSPACES CLASSIFICATION

#### A. Classification criteria

The classification is conducted on the basis of the topology of the singular curves in the workspace, which we characterize by (*i*) the number of cusps and (*ii*) the number of nodes or intersecting points. A cusp (resp. a node) is associated with one point with three equal IKS (resp. with two pairs of equal IKS). These singular points are interesting features for characterizing the workspace shape and the accessibility in the workspace.

#### B. Number of cusps

For now on and without loss of generality, $d_2$ is set to 1. Thus, we need handle only three parameters $d_3$, $d_4$ and $r_2$. Efficient computational algebraic tools were used in [10] to provide the equations of five separating surfaces, which were shown to divide the parameter space into 105 cells. But [11] showed that only 5 cells should exist, which means that one or more surfaces among the five ones found in [10] are not relevant. However, [11] did not try to find which surfaces are really separating. To derive the equations of the true separating surfaces, we need to investigate the transitions between the five cases. First, let us recall the five different cases found in [11]. The first case is a binary manipulator (i.e. it has only two IKS) with no cusp and a hole (Fig. 4). The remaining four cases are quaternary manipulators (i.e. with four IKS). The second case is a manipulator with four cusps on the internal boundary. Fig. 5 shows a manipulator of this case with a hole and two nodes. Note that the manipulator shown in Fig. 2 is another instance of case 2, although it has no node and no hole (see section *C*). Transition between case 1 and case 2 is a manipulator having a pair of points with four equal IKS, where two nodes and one cusp coincide [15].

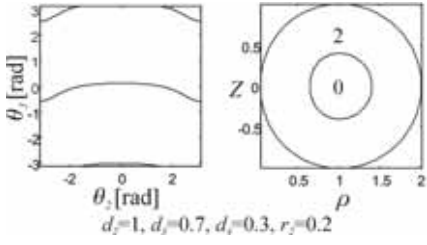

Figure 4. Manipulator of case 1.

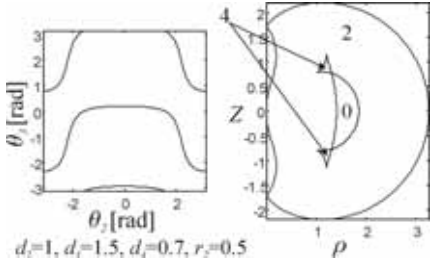

Figure 5. Manipulator of case 2.

Deriving the condition for the inverse kinematic polynomial to have four equal roots yields the equation of the separating surface [15]

$$d_4 = \sqrt{\frac{1}{2}\left(d_3^2 + r_2^2 - \frac{(d_3^2 + r_2^2)^2 - d_3^2 + r_2^2}{AB}\right)} \quad (2)$$

where

$$A = \sqrt{(d_3+1)^2 + r_2^2} \text{ and } B = \sqrt{(d_3-1)^2 + r_2^2}. \quad (3)$$

The third case is a manipulator with only two cusps on the internal boundary, which looks like a fish with one tail (Fig. 6). As shown in next section, an intermediate state exists between the manipulator shown in Fig. 5 and the one depicted in Fig. 6. This intermediate state is a variant of case 2 with two nodes and no hole (the upper and lower segments of the internal boundary cross, forming a '2-tail fish', see Fig. 11).

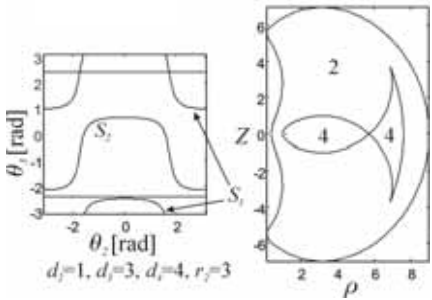

Figure 6. Manipulator of case 3.

As shown in [15], transition between case 2 and case 3 is characterized by a manipulator for which the singular line given by $\theta_3 = -\arccos(-d_3/d_4)$ is tangent to the singularity curve $S_1$. Expressing this condition yields the equation of the separating surface

$$d_4 = \frac{d_3}{1+d_3} \cdot A \quad (4)$$

where *A* is given by (3).

The fourth case is a manipulator with four cusps. Unlike case 2, the cusps are not located on the same boundary (Fig. 7).

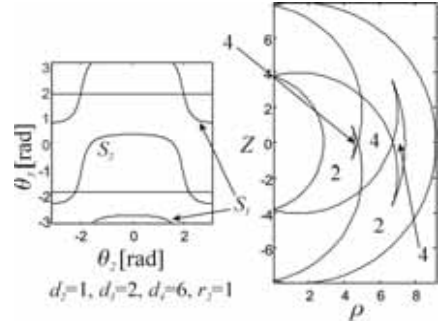

Figure 7. Manipulator of case 4.

Transition between case 3 and case 4 is characterized by a manipulator for which the singular line given by $\theta_3 = -\arccos(-d_3/d_4)$ is tangent to the singularity curve $S_2$ [15]. Expressing this condition yields the equation of the separating surface

$$d_4 = \frac{d_3}{d_3 - 1} \cdot B \text{ and } d_3 > 1 \quad (5)$$

where *B* is given by (3). As shown in next section, an intermediate state exists between the manipulator shown in Fig. 6 and the one depicted in Fig. 7. This intermediate state is a variant of case 3, which features two additional nodes that result from the intersection of the two workspace boundaries (like in Fig. 3).

Last case is a manipulator with no cusp. Unlike case 1, the internal boundary does not bound a hole but a region with 4 IKS. The two isolated singular points inside the inner region are associated with the two singularity lines.

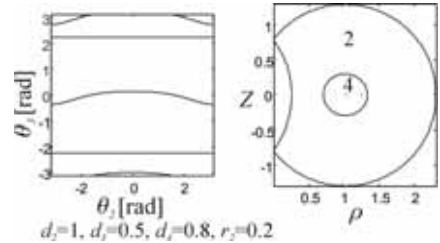

Figure 8. Manipulator of case 5.

Transition between case 4 and case 5 is characterized by a manipulator for which the singular line given by $\theta_3 = +\arccos(-d_3/d_4)$ is tangent to the singularity curve $S_1$ [15]. Expressing this condition yields the equation of the separating surface

$$d_4 = \frac{d_3}{1 - d_3} \cdot B \text{ and } d_3 < 1 \quad (6)$$

We have provided the equations of four surfaces that divide the parameters space into five domains where the number of cusps is constant. Fig. 9 shows the plots of these surfaces in a section ($d_3$, $d_4$) of the parameter space for $r_2=1$. Domains 1, 2, 3, 4 and 5 are associated with manipulators of case 1, 2, 3, 4 and 5, respectively. $C_1$, $C_2$, $C_3$ and $C_4$ are the right hand side of

(2), (4), (5) and (6), respectively.

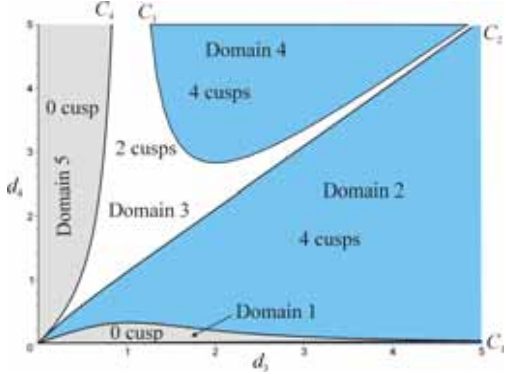

Figure 9.  Plots of the four separating surfaces in a section ($d_3$, $d_4$) of the parameter space for $r_2$=1.

It is interesting to see the correspondence between the equations found with pure algebraic reasoning in [10] and those provided in this paper. The five equations found in [10] are

$$-d_3 + d_4 r_2^2 + d_4 = 0 \quad (7)$$

$$d_3^2 - d_4^2 + r_2^2 = 0 \quad (8)$$

$$d_4^2 d_3^6 - d_4^4 d_3^4 + 3 d_4^2 d_3^4 r_2^2 - 2 d_4^2 d_3^3 + 2 d_4^4 d_3^2 - 2 d_4^4 d_3^2 r_2^2 \\ + d_4^2 d_3^2 + 3 d_4^2 d_3^2 r_2^4 - d_3^2 r_2^2 - 2 d_4^4 r_2^2 - d_4^4 r_2^4 - d_4^4 + d_4^2 r_2^6 \quad (9) \\ + d_4^2 r_2^2 + 2 d_4^2 r_2^4 = 0$$

$$d_3^2 r_2^2 + d_3^2 - 2 d_3^3 + d_3^4 - d_4^2 + 2 d_3 d_4^2 - d_3^2 d_4^2 = 0 \quad (10)$$

$$d_3^2 r_2^2 + d_3^2 + 2 d_3^3 + d_3^4 - d_4^2 - 2 d_3 d_4^2 - d_3^2 d_4^2 = 0 \quad (11)$$

Equation (9) is a second-degree polynomial in $d_4^2$. Solving this quadratics for $d_4$ shows that (9) can be rewritten as

$$d_4 = \sqrt{\frac{1}{2}\left(d_3^2 + r_2^2 - \frac{(d_3^2 + r_2^2)^2 - d_3^2 + r_2^2}{AB}\right)} \text{ or }$$

$$d_4 = \sqrt{\frac{1}{2}\left(d_3^2 + r_2^2 + \frac{(d_3^2 + r_2^2)^2 - d_3^2 + r_2^2}{AB}\right)}$$

where $A$ and $B$ are defined in (3). The first branch is the separating surface $d_4=C_1$ between domains 1 and 2.

Equation (10) is a second-degree polynomial in $d_4$. By solving this quadratics for $d_4$ and assuming strictly positive values for $d_4$ and $r_2$, (10) can be rewritten as

$$(d_4 = \frac{d_3}{d_3 - 1} \cdot B \text{ and } d_3 > 1) \text{ or } (d_4 = \frac{d_3}{1 - d_3} \cdot B \text{ and } d_3 < 1)$$

where $B$ is defined in (3). These two branches are the separating surfaces $d_4=C_3$ and $d_4=C_4$, respectively.

In the same way, (11) can be rewritten as,

$$d_4 = \frac{d_3}{1 + d_3} \cdot A$$

which is the separating surface $d_4=C_2$.

Thus, (7) and (8) found in [10] do not define separating surfaces, and only one branch of (9) defines a separating surface.

### C. Number of nodes

In this section, we investigate each domain according to the number of nodes in the workspace.

#### 1) Domain 1

Since all manipulators in this domain are binary, they cannot have any node in their workspace. Thus, all manipulators in domain 1 have the same workspace topology, namely, 0 node, 0 cusp and a hole inside their workspace. This workspace topology is referred to as $WT_1$.

#### 2) Domain 2

Figures 5 and 2 show two distinct workspace topologies of manipulators in domain 2, which feature 2 nodes and 0 node and which we call $WT_2$ and $WT_3$, respectively. Transition between these two workspace topologies is one such that the two lateral segments of the internal boundary meet tangentially (Fig. 10).

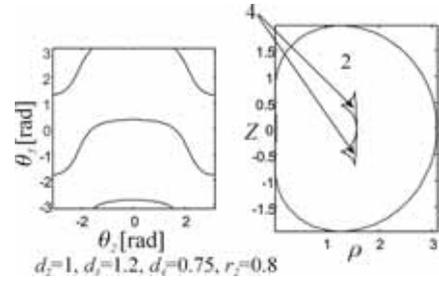

$d_2$=1, $d_3$=1.2, $d_4$=0.75, $r_2$=0.8

Figure 10.  Transition between $WT_2$ and $WT_3$.

Equation of this transition can be derived geometrically and the following equation is found [15]

$$d_4 = \frac{1}{2}(A - B) \quad (12)$$

where $A$ and $B$ are defined in (3).

As noted in section B, a third topology exists in this domain, where the internal boundary exhibits a '2-tail fish'. This workspace topology, which we call $WT_4$, features two nodes like in Fig. 5, but these nodes do not play the same role. They coincide with two isolated singular points, which are associated with the two singularity lines defined by $\theta_3=\pm\arccos(-d_3/d_4)$ (the operation point lies on the second joint axis and the inverse kinematics admits infinitely many solutions). Also, the nodes do not bound a hole like in Fig. 5 but a region with four IKS (Fig. 10).

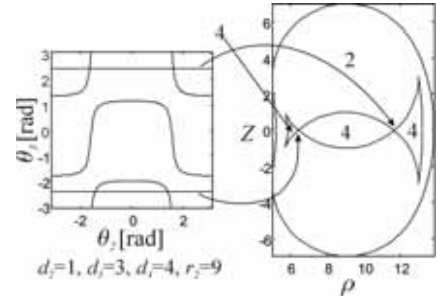

$d_2$=1, $d_3$=3, $d_4$=4, $r_2$=9

Figure 11.  Workspace topology $WT_4$.

Transition between $WT_3$ and $WT_4$ is a workspace topology such that the upper and lower segments of the internal boundary meet tangentially (Fig. 12).

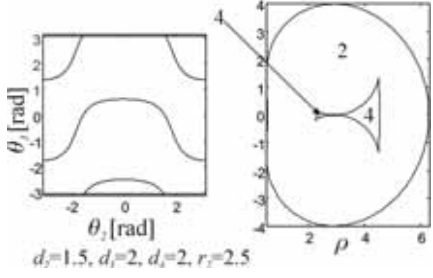

Figure 12. Transition between $WT_3$ and $WT_4$.

As shown in [15], this transition is the occurrence of the additional singularity $d_3 + c_3 d_4 = 0$, that is

$$d_4 = d_3 \qquad (13)$$

*3) Domains 3 and 5*

The internal boundary has either 2 cusp (domain 3) or 0 cusp (domain 5). This boundary is either fully inside the external boundary (like in Figs 6 and 8), or it can cross the external boundary, yielding two nodes as in Fig. 3 and 13. Thus, domain 3 (resp. domain 5) contains two distinct workspace topologies, which we call $WT_5$ (1 node) and $WT_6$ (resp. $WT_8$ and $WT_9$).

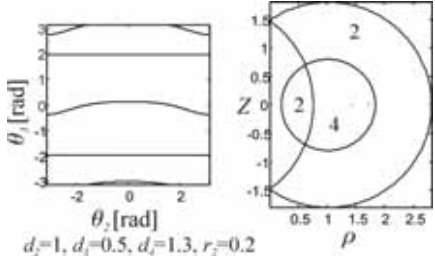

Figure 13. Workspace topology $WT_9$.

Transition between $WT_5$ and $WT_6$ and transition between $WT_8$ and $WT_9$ are such that the internal boundary meets the external boundary tangentially (Fig. 14).

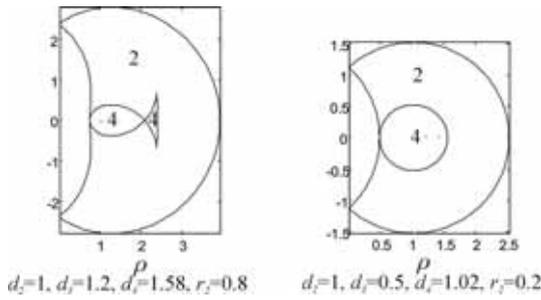

Figure 14. Transition between $WT_5$ and $WT_6$ (left) and between $WT_8$ and $WT_9$ right).

This transition can be derived geometrically and the following equation is found [15]

$$d_4 = \frac{1}{2}(A + B) \qquad (14)$$

where $A$ and $B$ are defined in (3).

*4) Domains 4*

Manipulators in domain 4 have four cusps and four nodes.

No subcase exist in this domain [15]. Such topologies are referred to as $WT_7$.

IV. RESULTS SYNTHESIS

*A. Parameter space partition*

Taking into account the nodes in the classification results in a new partition of the parameter space, as shown in Fig. 15, where $E_1$, $E_2$ and $E_3$ are the right hand side of (12), (13) and (14), respectively. Figure 15 depicts a section $(d_3, d_4)$ of the parameter space for $r_2=1$.

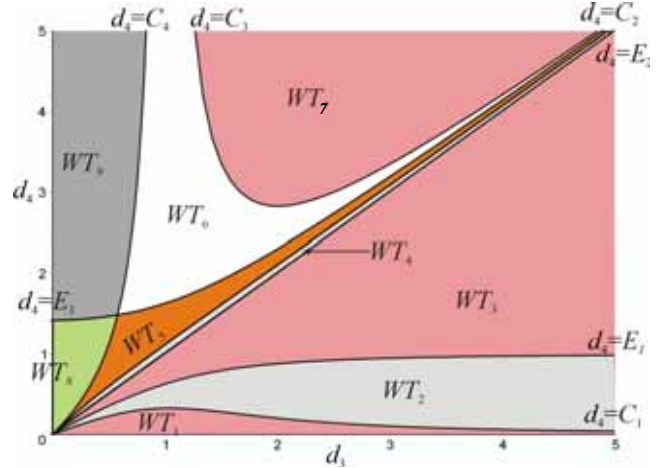

Figure 15. Parameter space partition according to the number of cusps and nodes (in a section $r_2=1$).

Plots of the separating surfaces in sections for different values of $r_2$ are shown in Fig. 16.

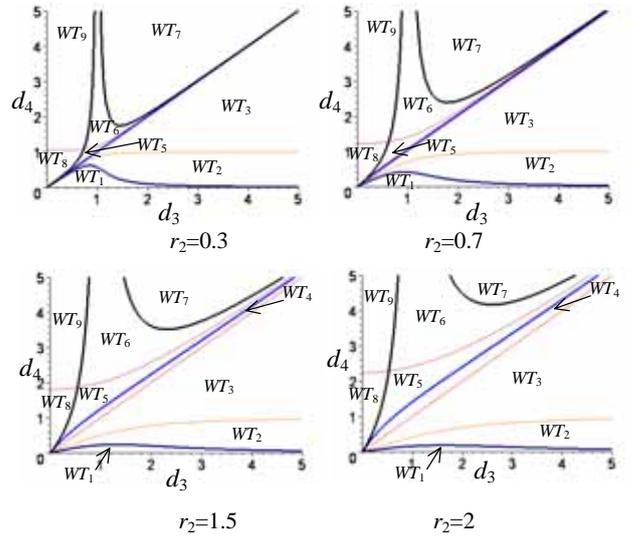

Figure 16. Separating surfaces for different values of $r_2$.

The areas associated with $WT_1$, $WT_2$, $WT_7$ and $WT_9$ decrease when $r_2$ increases. The area associated with $WT_4$ is very tiny, especially for small values of $r_2$. This means that few manipulators have a topology of the $WT_4$ type.

## B. Classification tree

A multi-level classification of the 3R orthogonal manipulators under study can be established by the classification tree shown in Fig. 17. For more legibility, only the generic cases are reported on this tree (i.e. manipulators on the separating surfaces of the parameter space are not reported). The root of the tree is the set of all manipulators under study and each leave is the set of manipulators with a completely specified workspace topology. The first level of the classification tree shows that a 3R orthogonal manipulator has either 2 aspects (if $d_3>d_4$), or it is quaternary and has no hole in its workspace (if $d_3<d_4$). The second level shows that (i) a 3R orthogonal manipulator with 2 aspects is either quaternary with 4 cusps (if $d_4>C_1$), or binary with no cusp, no node and a hole (if $d_4<C_1$) and (ii) a 3R orthogonal quaternary manipulator may have 4 cusps and 6 aspects (if $d_4>C_3$ or $d_4<C_2$), or 2 cusps and 5 aspects (if $C_2<d_4<C_3$ and $d_4<C_4$), or 0 cusp and 4 aspects (if $d_4>C_4$).

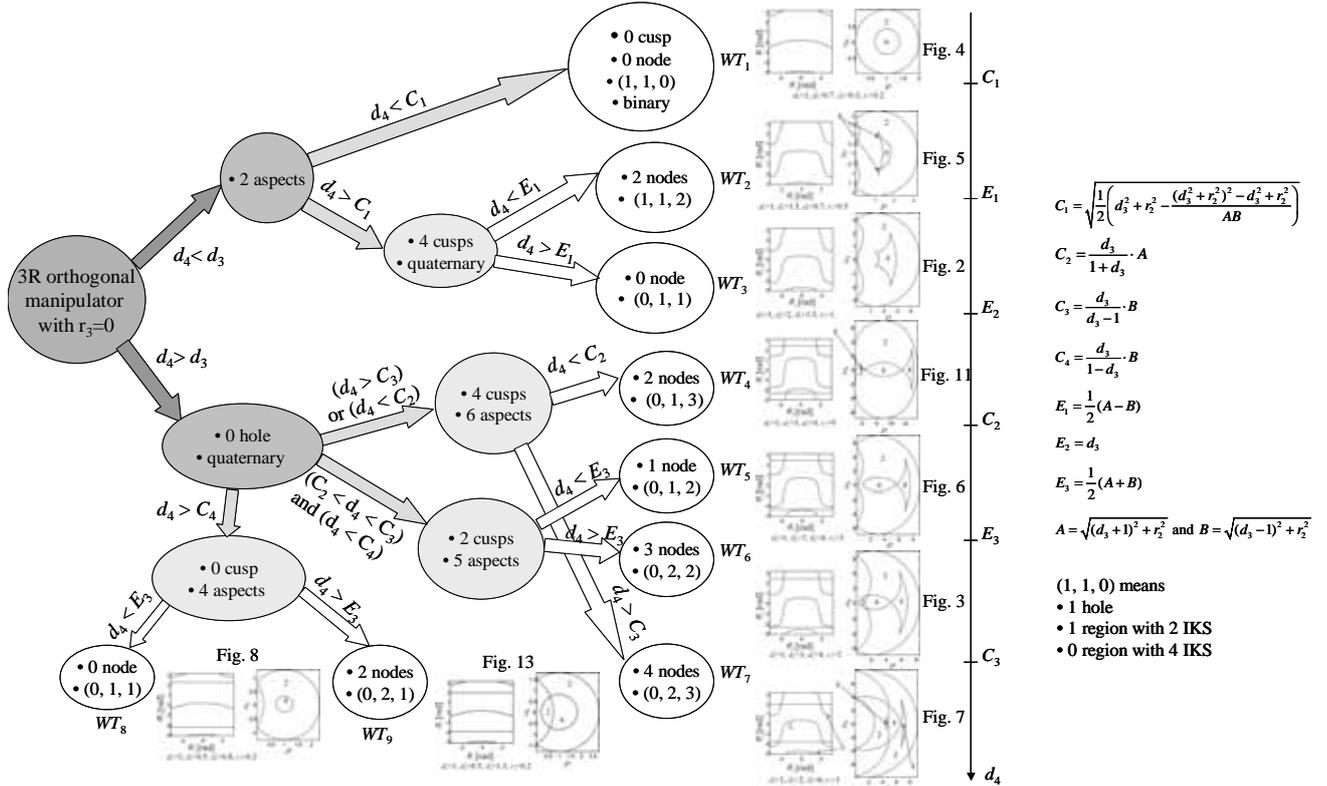

Figure 17. Classification tree.

## V. CONCLUSIONS

A family of 3R manipulators was classified according to the topology of the workspace, which was defined as the number of cusps and nodes. The design parameters space was shown to be divided into nine domains of distinct workspace topologies. Each separating surface was given as an explicit expression in the DH-parameters. Further work will investigate each domain according to various interesting design criteria.

## REFERENCES

[1] C.V. Parenti and C. Innocenti, "Position Analysis of Robot Manipulators: Regions and Sub-regions," in *Proc. Int. Conf. on Advances in Robot Kinematics*, pp 150-158, 1988.
[2] J. W. Burdick, "Kinematic analysis and design of redundant manipulators," PhD Dissertation, Stanford, 1988.
[3] P. Borrel and A. Liegeois, "A study of manipulator inverse kinematic solutions with application to trajectory planning and workspace determination," in *Proc. IEEE Int. Conf. Rob. and Aut.*, pp 1180-1185, 1986.
[4] J. W. Burdick, "A classification of 3R regional manipulator singularities and geometries," *Mechanisms and Machine Theory*, Vol 30(1), pp 71-89, 1995.
[5] P. Wenger, "Design of cuspidal and noncuspidal manipulators," in *Proc. IEEE Int. Conf. on Rob. and Aut.*, pp 2172-2177., 1997
[6] J. El Omri and P. Wenger, "How to recognize simply a non-singular posture changing 3-DOF manipulator," *Proc. 7th Int. Conf. on Advanced Robotics*, p. 215-222, 1995.
[7] V.I. Arnold, *Singularity Theory*, Cambridge University Press, Cambridge, 1981.
[8] P. Wenger, "Classification of 3R positioning manipulators," *ASME Journal of Mechanical Design*, Vol. 120(2), pp 327-332, 1998.
[9] P. Wenger, "Some guidelines for the kinematic design of new Manipulators," *Mechanisms and Machine Theory*, Vol 35(3), pp 437-449, 1999.
[10] S. Corvez and F. Rouiller,"Using computer algebra tools to classify serial manipulators,"in *Proc. Fourth International Workshop on Automated Deduction in Geometry*, Linz, 2002.
[11] M. Baili, P. Wenger and D. Chablat, "Classification of one family of 3R positioning manipulators, "in *Proc. 11th Int. Conf. on Adv. Rob.*, 2003.
[12] J. El Omri, 1996, "Kinematic analysis of robotic manipulators," PhD Thesis, University of Nantes (*in french*).
[13] D. Kohli and M. S. Hsu, "The Jacobian analysis of workspaces of mechanical manipulators," *Mechanisms an Machine Theory*, Vol. 22(3), p. 265-275, 1987.
[14] M. Ceccarelli, "A formulation for the workspace boundary of general n-revolute manipulators," *Mechanisms and Machine Theory*, Vol 31, pp 637-646, 1996.
[15] M. Baili, "Classification of 3R Orthogonal positioning manipulators, " technical report, University of Nantes, September 2003.